\documentclass[11pt,a4paper]{article}
\usepackage[hyperref]{naaclhlt2019}
\usepackage{times}
\usepackage{latexsym}
\usepackage{amssymb}

\usepackage{hhline}
\usepackage{tikz}
\usepackage{wrapfig}

\usepackage{todonotes}
\usepackage{subcaption}

\usepackage{url}
\usepackage{multirow}
\usepackage{tabularx}

\aclfinalcopy % Uncomment this line for the final submission
\def\aclpaperid{1874} %  Enter the acl Paper ID here

\makeatletter

\def\thickhline{%
  \noalign{\ifnum0=`}\fi\hrule \@height \thickarrayrulewidth \futurelet
   \reserved@a\@xthickhline}
\def\@xthickhline{\ifx\reserved@a\thickhline
               \vskip\doublerulesep
               \vskip-\thickarrayrulewidth
             \fi
      \ifnum0=`{\fi}}

\makeatother

\newlength{\thickarrayrulewidth}
\newcommand*\rot{\rotatebox{50}}

\title{Evaluating Rewards for Question Generation Models}

\author{Tom Hosking  \\
University College London \\
\tt{thomas.hosking.17@ucl.ac.uk} \\ \And
Sebastian Riedel\\
University College London \\
\tt{sriedel@ucl.ac.uk} \\
}

\date{}

\begin{document}
\maketitle
\begin{abstract}
  Recent approaches to question generation have used modifications to a Seq2Seq architecture inspired by advances in machine translation. Models are trained using teacher forcing to optimise only the one-step-ahead prediction.  However, at test time, the model is asked to generate a whole sequence, causing errors to propagate through the generation process (exposure bias). A number of authors have suggested that optimising for rewards less tightly coupled to the training data might counter this mismatch.
 We therefore optimise directly for various objectives beyond simply replicating the ground truth questions, including a novel approach using an adversarial discriminator that seeks to generate questions that are indistinguishable from real examples. We confirm that training with policy gradient methods leads to increases in the metrics used as rewards. We perform a human evaluation, and show that although these metrics have previously been assumed to be good proxies for question quality, they are poorly aligned with human judgement and the model simply learns to exploit the weaknesses of the reward source.
\end{abstract}

\section{Introduction}

\begin{table*}[t]
\def\arraystretch{1.25}
 \footnotesize
 \centering
 \begin{tabularx}{\textwidth}{ p{0.3\textwidth} p{0.649\textwidth} }
% ix=2005 (old)
% ix=5069
 \thickhline
  
  \multicolumn{2}{p{0.95\textwidth}}{
  \textbf{Context}
  }\\
  \multicolumn{2}{p{0.95\textwidth}}{
    although united methodist practices and interpretation of beliefs have evolved over time , these practices and beliefs can be traced to the writings of the church 's founders , especially \textbf{john wesley and charles wesley} ( anglicans ) , but also philip william otterbein and martin boehm ( united brethren ) , and jacob albright ( evangelical association ) .
    % Test
    } \\
    % \hline
%  \multicolumn{2}{p{\textwidth}}{
%  \textbf{Answer} &
% %  }\\
% %   \multicolumn{2}{p{\textwidth}}{
%   john wesley and charles wesley  \\
%   }\\
    % \thickhline
    \hhline{==}
  \textbf{Rewards} & \textbf{Output} \\ \hline
%   \hline

 {Ground Truth Question} &
   who were two of the founders of the united methodist church ?  \\
   \hline
 {No fine tuning} &
   which two methodist can be traced to the church 's founders ?  \\ \hline
 {LM} &
   according to the writings of the church 's founders , according to the writings of the church 's founders , [...]  \\ \hline
 {QA} &
   who in anglicans ?  \\ \hline
 {LM and QA} &
   who are the writings of the church 's founders ?  \\ \hline
 {Discriminator} &
   who founded the church 's founders ?  \\ \hline
 {Adversarial discriminator} &
   who were two western methodist practices ?  \\ \hline
 {LM, QA and adversarial discriminator} &
   who are the anglicans of the church ?  \\
 
\thickhline
 \end{tabularx}
 \caption{Example generated questions for various fine-tuning objectives. The answer is highlighted in bold. The model trained on a QA reward has learned to simply point at the answer and exploit the QA model, while the model trained on a language model objective has learned to repeat common phrase templates.}
 \label{tab:rlexamplecomparison-2}
 \end{table*}

Posing questions about a document in natural language is a crucial aspect of the effort to automatically process natural language data, enabling machines to ask clarification questions~\cite{quarc:emnlp18}, become more robust to queries~\cite{Yu2018}, and to act as automatic tutors~\cite{Heilman2010}.

% While questions often have a unique answer, the inverse is rarely true; there are multiple ways of phrasing a question, and there can be semantically different questions with the same answer. The ability to generate questions therefore provides a mechanism for augmenting existing question answering datasets or automatically annotating new ones, improving the resilience of question answering models.

Recent approaches to question generation have used Seq2Seq~\cite{Sutskever2014} models with attention~\cite{Bahdanau2014} and a form of copy mechanism~\cite{Vinyals2015, Gulcehre2016}. Such models are trained to generate a plausible question, conditioned on an input document and answer span within that document~\cite{Zhou2018,Du,Du2018,Maluuba}.

There are currently no dedicated question generation datasets, and authors have used the context-question-answer triples available in SQuAD. Only a single question is available for each context-answer pair, and models are trained using teacher forcing~\cite{TeacherForce}. This lack of diverse training data combined with the one-step-ahead training procedure exacerbates the problem of exposure bias~\cite{Ranzato2015}. The model does not learn how to distribute probability mass over sequences that are valid but different to the ground truth; during inference, the model must predict the whole sequence, and may not be robust to mistakes during decoding. 

Recent work has investigated training the models directly on a performance based objective, either by optimising for BLEU score~\cite{Kumar2018} or other quality metrics~\cite{Maluuba}. By decoupling the training procedure from the ground truth data, the model is able to explore the space of possible questions and learn to recover from suboptimal predictions during decoding. While the metrics used seem to be intuitively good choices, there is an assumption that they are good proxies for question quality which has not yet been confirmed.

Our contributions are as follows. We perform fine tuning using a range of rewards, including a novel adversarial objective that directly estimates the probability that a question was generated or came from the ground truth data. We show that although fine tuning leads to increases in reward scores, the resulting models perform worse when evaluated by human workers. We also demonstrate that the generated questions exploit weaknesses in the reward models.

\section{Background}

Many of the advances in natural language generation have been led by machine translation (MT)~\cite{Sutskever2014,Bahdanau2014,Gulcehre2016}.

Previous work on question generation has made extensive use of MT techniques. \citet{Du} use a Seq2Seq based model to generate questions conditioned on context-answer pairs, and build on this work by preprocessing the context to resolve coreferences and adding a pointer network~\cite{Du2018}. Similarly,~\citet{Zhou2018} use a part-of-speech tagger to augment the embedding vectors. Both authors perform a human evaluation of their models, and show significant improvement over their baseline. \citet{Kumar2018} use a similar model, but apply it to the task of generating questions without conditioning on a specific answer span. \citet{Song} use a modified context encoder based on multi-perspective context matching~\cite{Wang2016}.

\citet{Kumar} propose a framework for fine tuning using policy gradients and perform a human evaluation showing promising results. However, they use as rewards various similarity metrics that are still coupled to the ground truth. \citet{Maluuba} describe a Seq2Seq model with attention and a pointer network, with an additional encoding layer for the answer. They also describe a method for further tuning their model using policy gradients, with rewards given by an external language model and question answering (QA) system. Unfortunately they do not perform any human evaluation to determine whether this tuning led to improved question quality.

For the related task of summarisation,~\citet{Paulus2017} propose a framework for fine tuning a summarisation model using reinforcement learning, with the ROUGE similarity metric used as the reward.

\begin{table*}[ht!]
\footnotesize
\def\arraystretch{1.25}
\centering
\begin{tabular}{ c c c c | c c c c c}
\thickhline

\multicolumn{4}{c|}{\textbf{Features}} & \multicolumn{5}{c}{\textbf{Metrics}} \\
\rot{QA reward} & \rot{LM reward} & \rot{Discriminator reward} & \rot{Adversarial discriminator} &  \rot{NLL} & \rot{BLEU}  & \rot{QA} & \rot{LM} & \rot{Discriminator} \\
\hhline{====|=====}
%  \multicolumn{4}{c|}{No fine tuning} & \textbf{47.2} & 39.6 & \textbf{14.8}  & 66.2 & 64.7 & 5.0 \\
%  - & \checkmark & - & - & 44.8 & 38.9 & 12.9 & 62.5 & 51.3 & 6.5 \\
%  \checkmark & - & - & - & 39.3 & 42.3 & 9.3  & \textbf{70.1} & 292 & 10.4 \\
%  \checkmark & \checkmark & - & - & 45.7 & 39.1 & 12.2  & 68.2 & \textbf{48.4} & 7.9 \\
%  - & - & \checkmark & - & 45.6 & \textbf{38.8} & 13.0  & 64.1 & 55.3 & 7.5 \\
%  - & - & \checkmark & \checkmark & 44.8 & 46.0 & 12.1 & 63.7 & 65.2 & \textbf{15.8} \\
%  \checkmark & \checkmark & \checkmark & \checkmark & 45.1 & 40.6 & 12.4  & 67.5 & 58.0 & 15.0 \\
%  \multicolumn{4}{c|}{No fine tuning} & \textbf{47.2} & 39.6 & \textbf{14.8}  & 66.2 & 64.7 & 5.0 \\
 - & \checkmark & - & -  & -0.7 & -1.9 & -3.7 & -13.4 & +1.5 \\
 \checkmark & - & - & -  & +1.7 & -4.5  & \textbf{+3.9} & +226 & +5.4 \\
 \checkmark & \checkmark & - & -  & -0.5 & -2.6  & +2.0 & \textbf{-16.3} & +2.9 \\
 - & - & \checkmark & -  & \textbf{-0.8} & -1.8  & -2.1 & -9.4 & +2.5 \\
 - & - & \checkmark & \checkmark  & +6.4 & -2.7 & -2.5 & -1.0 & \textbf{+10.8} \\
 \checkmark & \checkmark & \checkmark & \checkmark  & +1.0 & -2.4  & +1.3 & -6.2 & +10.0 \\
\thickhline
 \end{tabular}
 \caption{Changes in automatic evaluation metrics after models were fine tuned on various objectives. QA refers to the F1 score obtained by a question answering system on the generated questions. LM refers to the perplexity of generated questions under a separate language model. The discriminator reward refers to the percentage of generated sequences that fooled the discriminator. Lower LM and NLL scores are better. BLEU scores decreased in all cases.}
 \label{tab:results-rl}
 \end{table*}

\begin{table*}[ht!]
\def\arraystretch{1.25}
\footnotesize
\centering
\begin{tabular}{ l | c c c}
\thickhline
Model & Fluency & Relevance  \\ 
 \hhline{=|===} 
% Baseline & 2.51 & 1.85 & 9.4 \\ 
No fine tuning & \textbf{3.34} & \textbf{3.12}  \\ 
% +Latent Switch +Latent Location & \textbf{3.51} & \textbf{3.42}  \\ 
+QA, LM rewards & 3.05 & 2.75  \\ 
+QA, LM, discriminator rewards +Adversarial discriminator & 2.89 & 2.82  \\ 
\hline
Ground Truth & 4.67 & 4.72 \\ 
\thickhline
\end{tabular}
\caption{Summary of human evaluation of selected models}
\label{tab:results-human}
\end{table*}

\begin{figure*}[ht]
\centering
    \begin{subfigure}[t]{0.49\textwidth}
        \centering
        \includegraphics[width=0.99\textwidth]{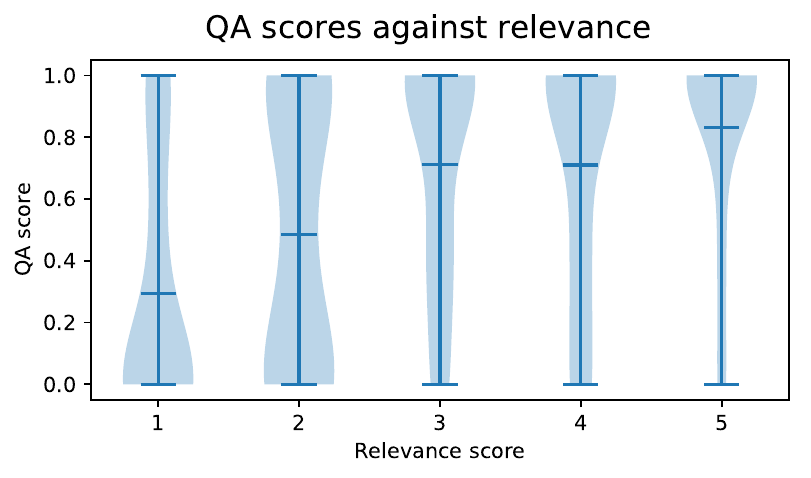}
        \caption{QA scores plotted against human relevance scores for all rated questions.}
    \end{subfigure}%
~
    \begin{subfigure}[t]{0.49\textwidth}
        \centering
        \includegraphics[width=0.99\textwidth]{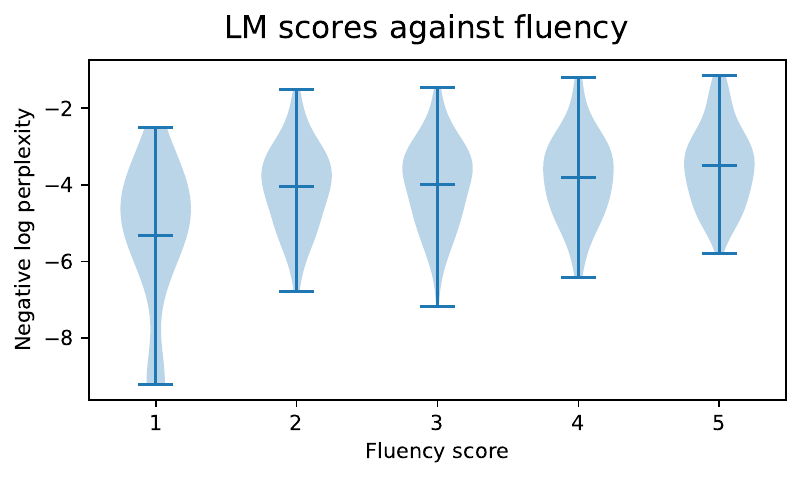}
        \caption{LM scores plotted against human fluency scores for all rated questions.}
    \end{subfigure}
\caption{Comparison of human and automatic metrics.}
\label{fig:human_qa_rel}
\end{figure*}

\section{Experimental setup}

The task is to generate a natural language question, conditioned on a document and the location of an answer within that document. For example, given the input document ``this paper investigates rewards for question generation" and answer ``question generation", the model should produce a question such as ``what is investigated in the paper?"

\subsection{Model description}

We use the model architecture described by~\citet{Maluuba}. Briefly, this is a Seq2Seq model~\cite{Sutskever2014} with attention~\cite{Bahdanau2014} and copy mechanism~\citep{Vinyals2015, Gulcehre2016}. \citet{Maluuba} also add an additional answer encoder layer, and initialise the decoder with a hidden state constructed from the final state of the encoder. Beam search~\cite{BeamSearch} is used to sample from the model at inference time. We train the model using maximum likelihood before fine tuning. Our implementation achieves a BLEU-4 score~\cite{Papineni} of 13.5 on the test set used by~\citet{Du}, before fine tuning.

\subsection{Fine tuning}

Generated questions should be formed of language that is both \textit{fluent} and \textit{relevant} to the context and answer. Following~\citep{Maluuba}, we perform fine tuning on a trained model, using rewards given either by the negative perplexity under a LSTM language model, or the F1 score attained by a question answering (QA) system, or a weighted combination of both. The language model is a standard recurrent neural network formed of a single LSTM layer. For the QA system, we use QANet~\cite{Yu2018} as implemented by~\citet{QANetGithub}.

\subsection{Adversarial training}
% \todo{More weight on this? Needs more weight in conclusion}
Additionally, we propose a novel approach by learning the reward directly from the training data, using a \textit{discriminator} detailed in Appendix~\ref{sec:discriminator}. We generate questions for each context-answer pair in the training set using a generator trained by maximum likelihood, and train the discriminator to predict whether an input question was generated by our model, or originated from the training data. Keeping the discriminator fixed, we then fine-tune the generator, using as reward the probability estimated by the discriminator that a generated question was in fact real. In other words, the generator is rewarded for successfully fooling the discriminator. We also experiment with interleaving updates to the discriminator within the fine tuning phase, allowing the discriminator to become adversarial and adapt alongside the generator.

The rewards described above are used to update the model parameters via the REINFORCE policy gradient algorithm~\cite{Williams1992}. We teacher force the decoder with the generated sequence to reproduce the activations calculated during beam search, to enable backpropagation. All rewards are normalised with a simple form of PopArt~\cite{VanHasselt}, with the running mean $\mu_R$ and standard deviation $\sigma_R$ updated online during training. We continue to apply a maximum likelihood training objective during this fine tuning.

\subsection{Evaluation}

We report the negative log-likelihood (NLL) of the test set under the different models, as well as the corpus level BLEU-4 score~\cite{Papineni} of the generated questions compared to the ground truth. We also report the rewards achieved on the test set, as the QA, LM and discriminator scores.

% We report the macro-averaged F1 score attained by a separate question \textit{answering} system, as the QA score. This can be viewed as a form of reconstruction score, since it should be possible to recover the answer used to generate a good question. We also report the perplexity of generated questions under a LSTM language model (LM) trained on the questions from SQuAD.

For the human evaluation, we follow the standard approach in evaluating machine translation systems~\cite{Koehn2006}, as used for question generation by~\citet{Du2018}. We ask three workers to rate 300 generated questions between 1 (poor) and 5 (good) on two separate criteria: the fluency of the language used, and the relevance of the question to the context document and answer.

\section{Results}

Table~\ref{tab:results-rl} shows the changes in automatic metrics for models fine tuned on various combinations of rewards, compared to the model without tuning. In all cases, the BLEU score reduces, as the training objective is no longer closely coupled to the training data. In general, models achieve better scores on the metrics on which they were fine tuned. Jointly training on a QA \textit{and} LM reward results in better LM scores than training on only a LM reward; the LM score did not increase smoothly when used as the sole objective, and we believe the additional QA reward acts as a form of regularisation. We conclude that fine tuning using policy gradients can be used to attain higher rewards, as expected.

Table~\ref{tab:results-human} shows the human evaluation scores for a subset of the fine tuned models. The model fine tuned on a QA and LM objective is rated as significantly worse by human annotators, despite achieving higher scores in the automatic metrics. In other words, the training objective given by these reward sources does not correspond to true question quality, despite them being intuitively good choices.

The model fine tuned using an adversarial discriminator has also failed to achieve better human ratings, with the discriminator model unable to learn a useful reward source. Although the training process was stable and robust to different initialisations, and the outputs do not appear to be significantly worse, we conclude that the discriminator was unable to learn a sufficiently useful distinction between generated and real questions, and the additional fine tuning procedure simply added unwanted noise to the model predictions.

Table~\ref{tab:rlexamplecomparison-2} shows an example where fine tuning has not only failed to improve the quality of generated questions, but has caused the model to exploit the reward source. The model fine tuned on a LM reward has degenerated into producing a loop of words that is evidently deemed probable, while the model trained on a QA reward has learned that it can simply point at the location of the answer. This observation is supported by the metrics; the model fine tuned on a QA reward has suffered a catastrophic worsening in LM score of +226.

Figure~\ref{fig:human_qa_rel} shows the automatic scores against human ratings for all rated questions. The correlation coefficient between human relevance and automatic QA scores was 0.439, and between fluency and LM score was only 0.355. While the automatic scores are good indicators of whether a question will achieve the lowest human rating or not, they do not differentiate clearly between the higher ratings: training a model on these objectives will not necessarily learn to generate better questions. A good question will likely attain a high QA and LM score, but the inverse is not true; a sequence may exploit the weaknesses of the metrics and achieve a high score \textit{despite} being unintelligible to a human. We conclude that fine tuning a question generation model on these rewards does not lead to better quality questions.

\section{Conclusion}

In this paper, we investigate the use of external reward sources for fine tuning question generation models to counteract the lack of task-specific training data. We show that although fine tuning can be used to attain higher rewards, this does not equate to better quality questions when rated by humans. Using QA and LM rewards as a training objective causes the generator to expose the weaknesses in these models, which in turn suggests a possible use of this approach for generating adversarial training examples for QA models. The QA and LM scores are well correlated with human ratings at the lower end of the scale, suggesting they could successfully be used as part of a reranking or filtering system. We plan to research overgenerating questions and using the reward signals to rerank the outputs, thereby including the inductive bias the rewards represent without allowing the model to exploit them.

\bibliography{naaclhlt2019}
\bibliographystyle{acl_natbib}

\appendix

\section{Discriminator architecture}
\label{sec:discriminator}

\begin{figure}[ht]
\centering
\includegraphics[width=0.5\textwidth]{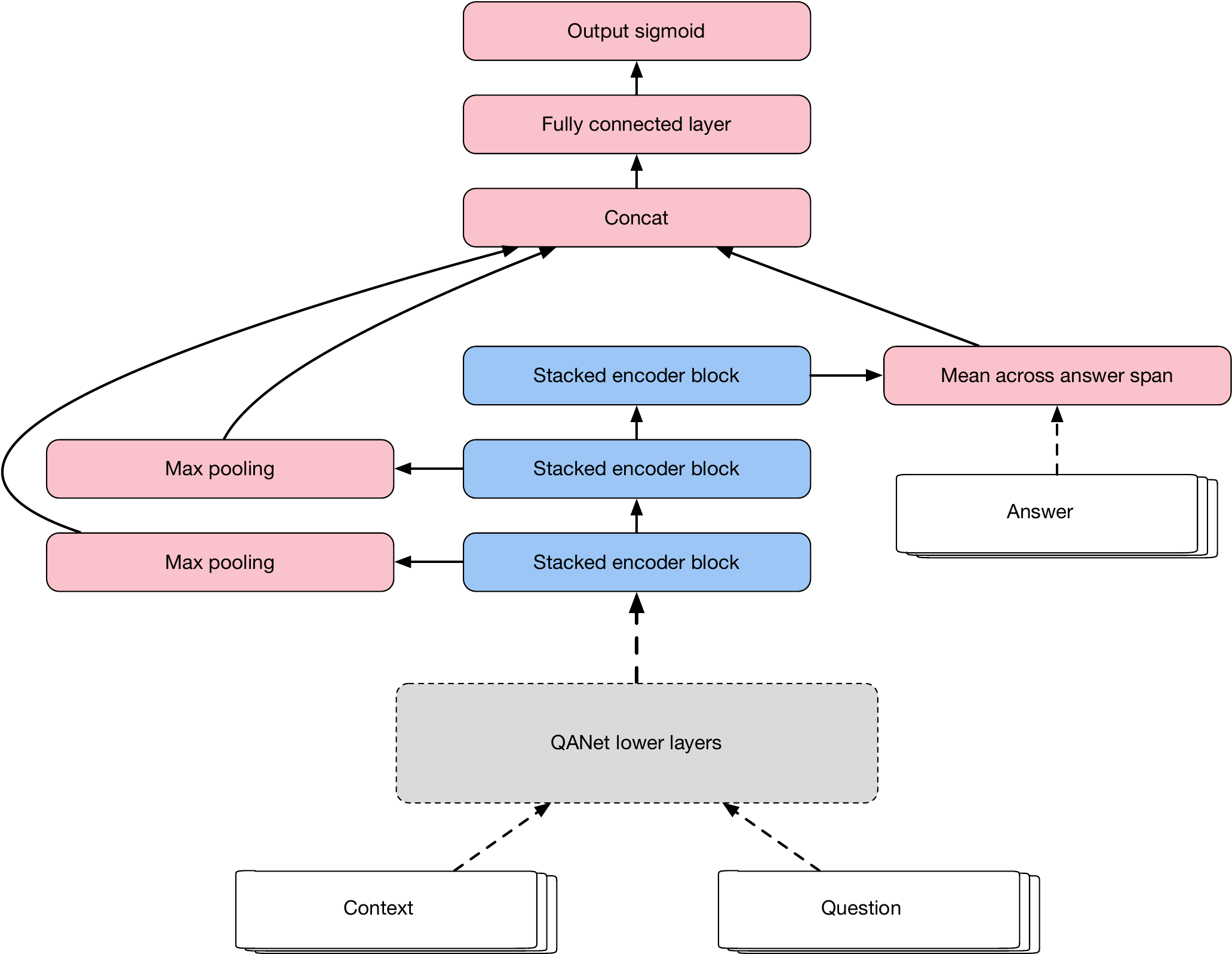}
\caption{Discriminator architecture diagram.}
\label{fig:disc_arch}
\end{figure}

We use an architecture based on a modified QANet as shown in Figure~\ref{fig:disc_arch}, replacing the output layers of the model to produce a single probability. Since the discriminator is also able to consider a full context-question-answer triple as input (as opposed to a context-question pair for the QA task), we fuse this information in the output layers.

Specifically, we apply max pooling over time to the output of the first two encoders, and we took the mean of the outputs of the third encoder that formed part of the answer span. These three reduced encodings were concatenated, a $64$ unit hidden layer with ReLU activation applied, and the output passed through a single unit sigmoid output layer to give the estimated probability that an input context-question-answer triple originated from the ground truth dataset or was generated.

\end{document}